\pdfoutput=1
\documentclass[11pt]{article}

\usepackage[final]{acl}

\usepackage{times}
\usepackage{latexsym}

\usepackage[T1]{fontenc}
\usepackage[utf8]{inputenc}

\usepackage{microtype}

\usepackage{inconsolata}

\usepackage{graphicx}
\usepackage{paralist}
\usepackage{enumitem}
\usepackage{amsmath}
\usepackage{amssymb} % For symbols like hearts, spades, etc.
\usepackage{pifont} % For symbols like hearts, spades, etc.
\usepackage[hang,flushmargin]{footmisc}

\newcommand{\heart}{\ding{170}} % heart symbol
\newcommand{\spade}{\ding{171}} % spade symbol
\newcommand{\diamondshape}{\ding{169}} % diamond symbol
\newcommand{\club}{\ding{168}} % club symbol

\title{LoGU: Long-form Generation with Uncertainty Expressions}

\author{
Ruihan Yang\textsuperscript{\spade}\thanks{Equal contribution. Work done during the internship at Tencent AI Lab. \dag Corresponding authors.}, Caiqi Zhang\textsuperscript{\heart}\footnotemark[1], Zhisong Zhang\textsuperscript{\diamondshape\dag}, \\
\textbf{Xinting Huang}\textsuperscript{\diamondshape}, 
\textbf{Sen Yang}\textsuperscript{\club}, 
\textbf{Nigel Collier}\textsuperscript{\heart},
\textbf{Dong Yu}\textsuperscript{\diamondshape}, 
\textbf{Deqing Yang}\textsuperscript{\spade\dag} \\
\textsuperscript{\spade}Fudan University,
\textsuperscript{\heart}University of Cambridge,
\textsuperscript{\diamondshape}Tencent AI Lab, \\
\textsuperscript{\club}The Chinese University of Hong Kong \\
{\small \texttt{\{rhyang17, deqingyang\}@fudan.edu.cn, cz391@cam.ac.uk, zhisonzhang@tencent.com}}
}

\usepackage{microtype}
\usepackage{graphicx}
\usepackage{arydshln}
\usepackage{inconsolata}
\usepackage{todonotes}
\usepackage{tabularx}
\usepackage{booktabs} % for better-looking tables
\usepackage{multirow}
\usepackage{amsmath}
\usepackage{amssymb}
\usepackage{graphicx}
\usepackage{subcaption}
\usepackage{fge}
\usepackage{colortbl}

\usepackage[framemethod=TikZ]{mdframed} 
\usepackage{listings}
\usepackage{longtable}
\usepackage{tcolorbox}
\usepackage{xspace}
\usepackage{wrapfig}
\usepackage{xcolor}
\usepackage{pgfplots}

\newcommand{\rparagraph}[1]{\vspace{1.2mm}\noindent\textbf{#1.}}

\definecolor{Gray}{gray}{0.92}
\definecolor{racing-green}{rgb}{0.0, 0.8, 0.6}
\definecolor{awesome-red}{rgb}{1.0, 0.13, 0.32}
\definecolor{LightCyan}{rgb}{0.88,1,1}
\definecolor{darkgreen}{RGB}{0,150,0}
\definecolor{Ground}{RGB}{255,184,55}
\definecolor{Dirt}{RGB}{191,169,115}
\definecolor{Pink}{RGB}{226,184,176}
\definecolor{Violet}{RGB}{163,148,170}

\newcommand{\ie}{\textit{i}.\textit{e}.,\ }
\newcommand{\eg}{\textit{e}.\textit{g}.,\ }

\newcolumntype{g}{>{\columncolor{Ground!7}}c}
\newcolumntype{d}{>{\columncolor{cyan!6}}c}
\newcolumntype{f}{>{\columncolor{lime!6}}c}
\newcolumntype{v}{>{\columncolor{purple!6}}c}

\newcommand{\sft}{\textsc{LoGU-SFT}\xspace}
\newcommand{\dpo}{\textsc{LoGU-DPO}\xspace}
\newcommand{\task}{\textsc{LoGU}\xspace}
\begin{document}
\maketitle

% \renewcommand{\thefootnote}{\dag}%
% \footnotetext{Corresponding authors.}

% Revert footnote numbering back to normal starting from 1
\renewcommand{\thefootnote}{\arabic{footnote}}
\setcounter{footnote}{0}

\begin{abstract}

While Large Language Models (LLMs) demonstrate impressive capabilities, they still struggle with hallucinations. A promising approach to mitigate hallucinations is enabling models to express uncertainty when unsure. Previous research on uncertainty estimation has primarily focused on short-form QA, but real-world applications often require much longer responses. In this work, we introduce the task of \textbf{Lo}ng-form \textbf{G}eneration with \textbf{U}ncertainty (LoGU), which requires the models to explicitly express uncertainty during the generation.
We identify two key challenges: \textit{Uncertainty Suppression}, where models hesitate to express uncertainty, and \textit{Uncertainty Misalignment}, where models convey uncertainty inaccurately. To tackle these challenges, we propose a novel decomposition-based data collection framework and a two-stage training pipeline. Specifically, we use supervised fine-tuning (SFT) for uncertainty suppression problem and direct preference optimization (DPO) for uncertainty misalignment. 
Experiments on three long-form datasets demonstrate the effectiveness of our approach, showing improvements in factual accuracy, reduction of incorrect statements, and preservation of the overall comprehensiveness of the generated responses. Further analysis reveals that baseline methods tend
to express uncertainty in vague and broad terms, while our method generates more specific and targeted uncertainty expressions.
\footnote{Project page: \url{https://github.com/rhyang2021/LoGU}.}

\end{abstract}

\section{Introduction}

While large language models (LLMs) demonstrate remarkable performance across various domains, they still suffer from a significant limitation: the generation of factually incorrect statements (\ie, hallucinations) \citep{zhao2023survey, hadi2023survey, chang2024survey, chen2024see}. This issue hinders the broader adoption of LLMs in real-world applications that demand highly reliable and accurate responses \citep{zhang-etal-2024-need, zhang2024luq, jiang-etal-2025-large}. 
Alleviating hallucinations remains a challenging and long-standing problem \citep{gekhman2024does, lin2024flame, he2025supposedly}. LLMs often produce responses even in the absence of sufficient knowledge \citep{xiong2023llms, kang2024unfamiliar}. 
Recognizing these intrinsic limitations, \textbf{it is essential for LLMs to explicitly express uncertainty during generation}.

\begin{figure}[t]
    \centering
    \includegraphics[width=0.99\columnwidth]{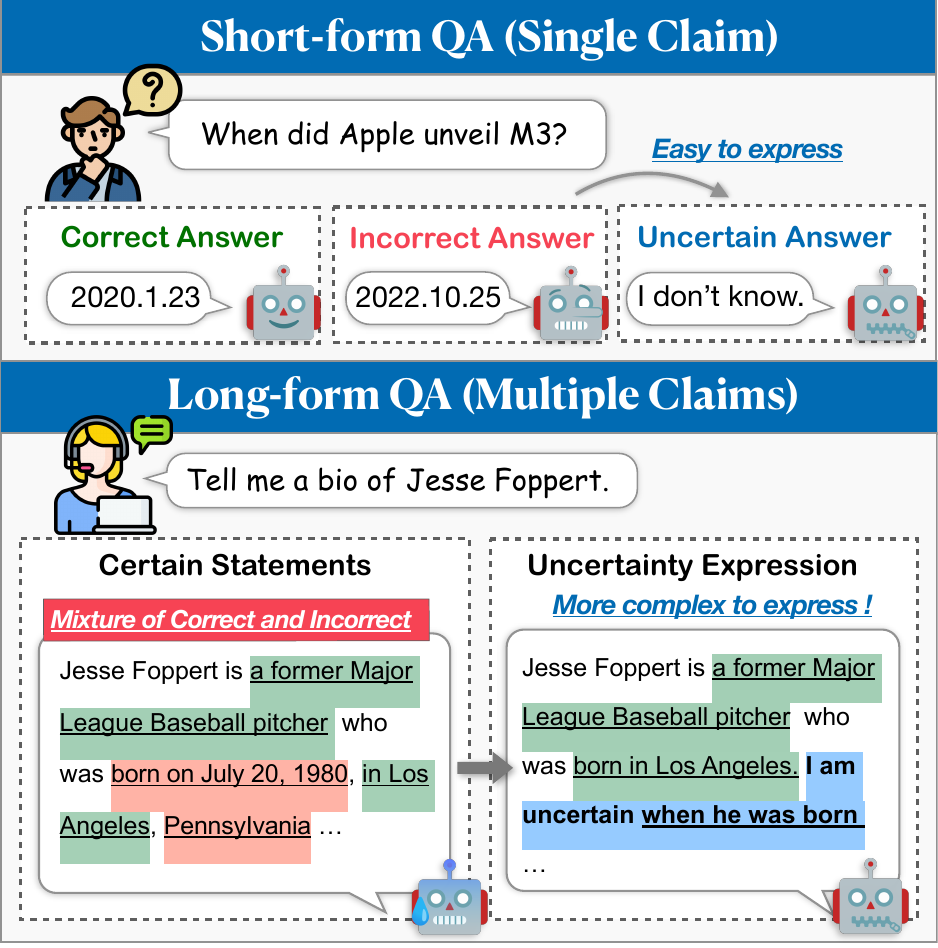}
    \caption{Existing works on short-form QA only give an overall uncertainty to the query. On the contrary, the long-form QA responses consist of both factually correct and incorrect claims, requiring more fine-grained uncertainty expressions.}
    \label{fig:problem}
\end{figure}

Previous work on enabling LLMs to express uncertainty has primarily focused on short-form question answering (QA) tasks \citep{lin2022teaching, tian-etal-2023-just, li2024know, zhang-etal-2024-r}. However, real-world applications often require much longer responses, sometimes spanning hundreds or even thousands of words \citep{bai2023longbench, zhang2024luq, yuan2024lv}. As shown in Figure~\ref{fig:problem}, unlike short-form responses, which typically contain a single claim, long-form responses involve multiple claims, increasing the complexity of accurately modeling uncertainty. 

In this paper, we introduce the task of \textbf{Lo}ng-form \textbf{G}eneration with \textbf{U}ncertainty (\task), which focuses on enabling LLMs to express uncertainty accurately in long-form responses. We emphasize linguistic expression of uncertainty rather than post-hoc estimation \citep{zhang2024luq, fadeeva-etal-2023-lm}. We argue that embedding uncertainty expressions directly in the generation is more human-interpretable and aligns more naturally with everyday communication.

We first identify two sub-challenges for \task: \textit{Uncertainty Suppression} and \textit{Uncertainty Misalignment}. \textit{Uncertainty Suppression} refers to the reluctance of LLMs to express uncertainty, while \textit{Uncertainty Misalignment} denotes cases where LLMs express uncertainty inaccurately. Ideally, LLMs should express uncertainty only when dealing with unknown facts.
To address these challenges, we propose a novel decomposition-based data collection framework and a two-stage training pipeline. 

\begin{figure*}[t!]
    \centering
    \includegraphics[width=0.96\textwidth]{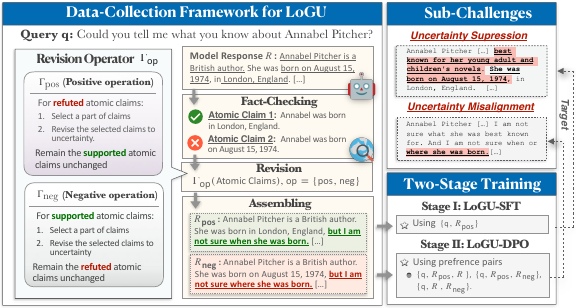}
    \caption{
    Our framework employs a decomposition-based approach: first, decompose responses into atomic claims for fact-checking and then revise selected atomic claims with uncertainty expressions. The revised atomic claims are assembled to a coherent response. The collected data is then used in the two-stage training pipeline: \sft and \dpo, which address \textit{uncertainty suppression} and \textit{uncertainty misalignment}, respectively.}
    \vspace{-2mm}
    \label{fig:main}
\end{figure*}

As illustrated in Figure~\ref{fig:main}, during data collection, we first decompose responses into atomic claims, then selectively revise them to include uncertainty expressions, and finally reassemble the atomic claims.
We use the collected data in a two-stage training pipeline consisting of \sft and \dpo. In \sft, we encourage the model to express uncertainty during generation to mitigate the \textit{uncertainty suppression} problem. In \dpo, response pairs with accurate and inaccurate uncertainty expressions are optimized using Direct Preference Optimization \citep{rafailov2024direct} to address the \textit{uncertainty misalignment}.

Experiments on three long-form datasets demonstrate the effectiveness of our approach, showing improvements in factual accuracy, reduction of incorrect statements, and preservation of the overall comprehensiveness of the generated responses. Further analysis reveals that baseline methods tend to express uncertainty in vague and broad terms, while our method generates more specific and targeted uncertainty expressions.

% Our main contributions are summarized as follows:

% \begin{itemize}[leftmargin=*]
%     \vspace{-2mm}
%     \item We introduce the task of \task, aiming to enable LLMs to accurately convey uncertainty in long-form generation.
%     \vspace{-2mm}
%     \item We identify two key challenges in \task: \textit{uncertainty suppression} and \textit{uncertainty misalignment}. To address these problems, we propose a refinement-based data collection framework and a two-stage training pipeline.
%     \vspace{-2mm}
%     \item Experiments on three long-form datasets demonstrate the effectiveness of our approach, showing improvements in factual accuracy, reducing incorrect statements, and preserving the overall comprehensiveness of the generated responses.
% \end{itemize}

\section{Task definition}
\label{sec:task}

\rparagraph{Definition} We define the \textbf{Lo}ng-form \textbf{G}eneration with \textbf{U}ncertainty (\task) task as follows. A language model, denoted by $\mathcal{M}$, is prompted with an open-ended query (\eg "Introduce \texttt{[a person]} to me.") to generate a long-form response \(R \sim \mathcal{M}(R \mid q)\) (\eg a biography). The response $R$ typically lacks explicit uncertainty expressions and includes both factually correct and incorrect statements. 
For fine-grained processing, we decompose the response \(R\) into \(N\) atomic claims \(C\), represented as \(C = \coprod_{i=1}^{N} c_i\). 
Each atomic claim is self-contained with only one piece of information.
These claims are categorized into three subsets: \(C_{\text{s}}\), containing claims supported by external knowledge; \(C_{\text{ns}}\), containing claims refuted by external knowledge; and \(C_\text{unc}\), including claims with uncertainty expressions. The \task task requires the model to reduce \(C_{\text{ns}}\) and express uncertainty in \(C_\text{unc}\) more accurately.

\rparagraph{Two Sub-Challenges} The \task task presents two main challenges:
\begin{inparaenum}[\it 1)]
    \item \textit{Uncertainty Suppression}: The model tends to avoid expressing uncertainty, displaying an overconfidence bias, which can result in reduced factual accuracy \citep{xiong2023can, tian-etal-2023-just, zhang2024luq};
    \item \textit{Uncertainty Misalignment}: The model's expressed uncertainty does not always align with its actual knowledge. It may show uncertainty even when it can provide correct answers.
\end{inparaenum}

\rparagraph{Primary Goals} Ideally, a model excelling in \task should maximize the following two aspects:

\begin{itemize}[leftmargin=*]
\item \textbf{Factual Accuracy (FA)} It represents the proportion of correctly generated deterministic claims in the model's output, reflecting how often the model generates factually correct statements. A higher FA score indicates that the model is generating a larger proportion of correct facts relative to incorrect ones. Specifically, FA is defined as:
\begin{equation*}
    FA = \frac{|C_{\text{s}}|}{|C_{\text{s}}| + |C_{\text{ns}}|}.
\end{equation*}

\item \textbf{Uncertain Accuracy (UA)} It represents the quality of uncertainty expressions by calculating how often the model accurately express uncertainty. Ideally, the model should express uncertainty in two specific cases:
(1) When the model does not have sufficient knowledge about a fact or is prone to make errors;
(2) When the uncertainty is expressed about specific, granular details (\eg ``I do not know \texttt{[a person]}'s birthday.''), rather than general or vague statements (\eg ``I am not sure about \texttt{[a person]}'s early life.'' is over broad). Within the uncertainty expression set \(C_\text{{unc}}\), we denote the truly uncertain claims as \(C_\text{{unc}}^\text{true}\), and UA is defined as:
\begin{equation*}
    \text{UA} = \frac{|C_\text{{unc}}^\text{true}|}{|C_\text{{unc}}|}.
\end{equation*}
\end{itemize}
More discussion on FA and UA are in Section \ref{sec:evaluation}.

\section{Methodology}
\label{sec:LoGU}

In this section, we explain our approach for the LoGU task in details. Figure \ref{fig:main} shows an overview of the data collection and our training pipeline.

\subsection{Data Collection Framework for \task}
\label{sec:datacollection}

For a target LLM \(\mathcal{M}\) and a list of queries $\{q_{i}\}_{i=1}^{n}$, we first gather an original set of query-response pairs, denoted as \(\mathcal{D} = \{(q_{1}, R_{1}), \ldots, (q_{n}, R_{n})\}\). Then we apply the following procedures to refine the responses to properly express uncertainty, with the help of an additional auxiliary LLM $\mathcal{A}$ (\eg GPT-4 \citep{achiam2023gpt}; See Appendix \ref{sec:logu_prompt} for the prompts of the procedures).

\rparagraph{Fact-Checking}
We use the \textsc{FActScore} \citep{min-etal-2023-factscore} for fact-checking. For each query $q$, the original response $R$ is decomposed into a series of atomic claims \(C\), with each claim containing only a single piece of information. This decomposition is performed by the auxiliary model $\mathcal{A}$. 
Next, we retrieve relevant passages from external sources (\eg Wikipedia or Google search) for each claim. The veracity of each claim is then assessed by model $\mathcal{A}$ based on its entailment by these passages. Claims that are supported are classified as \(C_{\text{s}}\), while those refuted are classified as \(C_{\text{ns}}\).

\rparagraph{Revision}
We apply the operator $\Gamma_{\texttt{op}}$ to the sets \(C_{\text{s}}\) and \(C_{\text{ns}}\), where \texttt{op} specifies the type of revision: either positive (\texttt{pos}) or negative (\texttt{neg}). Specifically, $\Gamma_{\texttt{op}}$ selects a subset of claims based on a given ratio and rewrite the selected claims into ones with uncertainty expressions. The details of $\Gamma_{\texttt{op}}$ are further explained in \S\ref{sec:operation}.

\rparagraph{Assembling}
After revising the atomic claims extracted from the response \(R\), we assemble these altered claims into a coherent and fluent refined response by prompting the auxiliary model $\mathcal{A}$. For the revised atomic claims \(\Gamma_{\texttt{op}}(C_{\text{s}})\) and \(\Gamma_{\texttt{op}}(C_{\text{ns}})\), the assembled response is denoted as \(R_{\texttt{op}}\). The final training dataset is represented as \(\mathcal{D}_{\texttt{op}} = \{(q_{1}, R_{\texttt{op}, 1}), \ldots, (q_{n}, R_{\texttt{op}, n})\}\).

\subsection{Revision Operator $\Gamma_{\texttt{op}}$}
\label{sec:operation}
Two types of operator $\Gamma_{\texttt{op}}$ (\ie \texttt{pos} and \texttt{neg}) are used to construct training data to address two specific problems: \textit{uncertainty suppression} and \textit{uncertainty misalignment}.

\rparagraph{Positive Revision Operation}
To address \textit{uncertainty suppression}, we construct QA pairs with explicitly expressed uncertainty. We first define $\Gamma_{\texttt{pos}}$, which revises refuted atomic claims ($C_{\text{ns}}$) into uncertainty expressions, while leaving supported claims ($C_{\text{s}}$) unchanged. We rewrite the atomic claims by prompting the auxiliary model $\mathcal{A}$ with ten manually designed distinct uncertainty expression patterns. For instance, the statement ``\texttt{[a person]} was born on August 15, 1974'' can be revised to ``I am not sure when \texttt{[a person]} was born.''

In real-world applications, the proportion of uncertainty expressions in long-form responses can influence the overall user experience. To prevent an overuse of uncertainty expressions from reducing readability, we apply a down-sampling mechanism to control their ratio. We therefore selectively revise a subset of refuted claims from $C_{\text{ns}}$ with a target uncertainty ratio $\alpha$ (more details about the impact of the uncertainty ratio in Appendix~\ref{sec:sft_thresholds}). Specifically, we select $\min\left(\frac{\alpha}{1-\alpha} |C_{\text{s}}|, |C_{\text{ns}}|\right)$ claims from $C_{\text{ns}}$, ensuring the final responses meet the desired uncertainty ratio $\alpha$. We denote the revised claims as $C_{\text{ns-unc}}$.

\rparagraph{Negative Revision Operation}
To address \textit{uncertainty misalignment}, where the model might express uncertainty about information it actually knows, we construct negative learning instances that contain such inaccurate uncertainty expressions. We define the operator $\Gamma_{\texttt{neg}}$ to revise supported atomic claims $C_{\text{s}}$ into uncertainty expressions while leaving refuted claims $C_{\text{ns}}$ unchanged. The output of $\Gamma_{\texttt{neg}}$ will be used as penalization signals to let the model not to inappropriately express uncertainty. Similar to $\Gamma_{\texttt{pos}}$, we also control the uncertainty ratio at $\alpha$. We select $\min\left(\frac{\alpha}{1-\alpha} |C_{\text{ns}}|, |C_{\text{s}}|\right)$ claims from $C_{\text{s}}$, which are then revised into uncertainty expressions, denoted as $C_{\text{s-unc}}$.

\rparagraph{Summary}
% For $\texttt{op} \in \{\texttt{pos}, \texttt{neg}\}$, the operator $\Gamma_{\texttt{op}}$ performs the following revisions to the original response $R$:
% \begin{equation*}
%     \vspace{-2pt}
%     \begin{aligned}
%         \Gamma_{\text{op}}(C_{\text{s}})&=
%         \begin{cases}
%         C_{\text{s}} & \text{if } \texttt{op}= \texttt{pos}\\
%         C_{\text{s}'\text{-unc}} & \text{if } \texttt{op} = \texttt{neg}
%         \end{cases} \\
%         \Gamma_{\text{op}}(C_{\text{ns}})&=
%         \begin{cases}
%         C_{\text{ns}'\text{-unc}} & \text{if }\texttt{op}=\texttt{pos}\\
%         C_{\text{ns}} & \text{if } \texttt{op} = \texttt{neg}
%         \end{cases}
%     \end{aligned}
% \end{equation*}
With the operations of $\Gamma_{\texttt{pos}}$ and $\Gamma_{\texttt{neg}}$, we can rewrite an original response $R$ into $R_{\texttt{pos}}$ and $R_{\texttt{neg}}$, respectively. $R_{\texttt{pos}}$ consists of the union of $C_{\text{s}}$ and $C_{\text{ns-unc}}$, while $R_{\texttt{neg}}$ includes $C_{\text{s-unc}}$ and $C_{\text{ns}}$. Note that since the down-sampling selection of atomic claims within $C_{\text{s}}$ and $C_{\text{ns}}$ is random, the revisions performed by $\Gamma_{\texttt{op}}$ on $C_{\text{s}}$ and $C_{\text{ns}}$ are not unique. This results in multiple possible variants, from which we select up to five samples per instance for our training process.

\subsection{Training Pipeline}
\label{sec:training}
With the constructed training data, we propose a two-stage training pipeline. In Stage 1 (\sft), our goal is to enable the model to express uncertainty, specifically addressing the problem of uncertainty suppression. In Stage 2 (\dpo), we aim to enhance the model's precision in expressing uncertainty, ensuring it avoids both exaggeration and understatement.

\rparagraph{\sft} We fine-tune the language model $\mathcal{M}$ using only positively refined responses ($R_{\texttt{pos}}$). Specifically, the model iteratively processes a sequence $t_1, t_2, \ldots, t_T$ consisting of queries and corresponding responses ($q$, $R_{\texttt{pos}}$). The objective is to minimize the cross-entropy loss $\mathcal{L}$:
\[
\mathcal{L} = -\frac{1}{T} \sum_{i=1}^T \log P(t_i \mid t_1, t_2, \ldots, t_{i-1}),
\]
where $P(t_i \mid t_1, t_2, \ldots, t_{i-1})$ denotes the probability of predicting token $t_i$ given all previous tokens. Importantly, the loss is computed exclusively for the response parts of the sequence, omitting the queries. After fine-tuning, we denote the model policy as $\pi_{\mathrm{sft}}$.

\rparagraph{\dpo} 
% Responses from $\mathcal{D}_{\texttt{pos}}$ are preferred over those from $\mathcal{D}$, which are in turn preferred over those from $\mathcal{D}_{\texttt{neg}}$. This prioritization results in $\binom{3}{2}$ preference pairs per query, forming the preference dataset $\mathcal{D}_{\texttt{pref}} = \{(q, y_w, y_l)\}$.
To refine the model's alignment in expressing uncertainty, we construct response pairs of good and unsatisfactory expressions from triplets\footnote{$R_{\texttt{pos}}$ is preferred over $R$, which is in turn preferred over $R_{\texttt{neg}}$. This results in $\binom{3}{2}$ preference pairs per query.} of ($R_{\texttt{pos}}$, $R$, $R_{\texttt{neg}}$). The negative refinements ($R_{\texttt{neg}}$) are used to address uncertainty misalignment by providing negative examples of inappropriate uncertainty expressions.

With these constructed response pairs ($q, y_w, y_l$), we employ the Direct Preference Optimization (DPO) algorithm~\citep{rafailov2024direct} for model training:
\begin{equation*}
\begin{aligned}
\mathcal{L}_\theta= & -\mathbb{E}_{\left(q, y_w, y_l\right) \sim \mathcal{D}}\left[\operatorname {log } \sigma \left(\beta \log \frac{\pi_\theta\left(y_w \mid q\right)}{\pi_{\text {sft }}\left(y_w \mid q\right)} \right. \right. \\
& \left.\left. -\beta \log \frac{\pi_\theta\left(y_l \mid q\right)}{\pi_{\text {sft }}\left(y_l \mid q\right)}\right)\right],
\end{aligned}
\end{equation*}
where $\pi_\theta$ is the model policy initialized from $\pi_{\text{sft}}$, $\beta$ controls the deviation from $\pi_{\text{sft}}$, and $\sigma$ represents the logistic function.
\section{Experiment Settings}

\subsection{Dataset and Models}
We use three datasets for long-form QA: (1) \textit{Bios} \citep{min-etal-2023-factscore}, which contains 500 individuals from Wikipedia with varying levels of popularity. The models are asked to generate biographies for the individuals. (2) \textit{LongFact} \citep{wei2024longfact}, which extends \textit{Bios} and includes 1,140 questions covering 38 manually-selected topics. (3) \textit{WildHallu} \citep{zhao2024wildhallu} comprises 7,917 entities derived from one million user-chatbot interactions in real-world settings. We also conduct experiments on an out-of-domain dataset ASQA \citep{stelmakh-etal-2022-asqa} in \S~\ref{sec:asqa_result}, to validate the generalization performance of our method.
Regarding the models, we use the Llama3-8B-Instruct \citep{llama3modelcard} and Mistral-7B-Instruct~\citep{jiang2023mistral}. For auxiliary model $\mathcal{A}$, we use GPT-4o. Implementation details are in Appendix \ref{sec:implementation_details}. 

\subsection{Evaluation Metrics}
\label{sec:evaluation}
We evaluate the performance of models in reflecting uncertainty in long-form generation using three core metrics: Factual Accuracy (FA), Uncertain Accuracy (UA), and number of incorrect claims (\#Incor). As discussed in \S \ref{sec:task}, the first two metrics are designed to measure the model's ability to accurately output factual statements and uncertainty expressions. In addition, we include \#Incor to explicitly measure the absolute number of errors that the model is making.

% \footnote{Recent studies show that models have better \textbf{discriminative ability} than \textbf{generative ability} \citep{saunders2022selfcritiquing}. Based on this, if the model is unable to answer a specific question correctly, it is very unlikely that the model can generate it correctly during the long-form generation.}

For Uncertain Accuracy, we approximate \(C_\text{{unc}}^\text{true}\) based on the following assumption: we regard a claim as truly uncertain if the model cannot directly answer it correctly in short-form QA.  Similar to previous work \citep{tian2023fine,zhang-etal-2024-self,farquhar2024detecting}, we first rewrite atomic facts into specific short-form questions. For example, the fact "\texttt{[a company]} was founded in 1996" is converted into the question "When was \texttt{[a company]} founded?". 
We present the short-form questions to the model and collect the model's answer. If the model fails to answer a question correctly, it indicates that the model have high uncertainty about that fact and the claim can be regarded as truly uncertain. We use the number of questions that lead to wrong answers as the approximation of \(C_\text{{unc}}^\text{true}\). Moreover, we focus on uncertainty about specific and granular details and exclude general and vague claims (as judged by GPT-4o; more discussion in Limitation Section).
% Questions that are specific and meet the criteria (2) for evaluating precise uncertainty are retained.
% (denoted as \(N_{\text{total}}\)).

% \paragraph{Incorrect Statements (\#Incor.):} 
In addition to FA and UA, we further include a metric of \#Incor., which tracks the total number of incorrect statements generated by the model. While accuracy focuses on the ratio of correct to incorrect facts, the \#Incor. metric directly measures the absolute number of incorrect statements. Reducing such incorrect facts is an essential goal for improving the factuality of long-form responses, as a high number of incorrect statements directly undermines the overall reliability of the model's output.

\begin{table*}[t!]
\setlength\tabcolsep{10pt}
\scalebox{0.8}[0.8]{ 
% \small
  \centering
    \begin{tabular}{llvvvdddggg}
    \toprule
     & \multirow{2}{*}{\textbf{Method}} & \multicolumn{3}{c}{\textbf{Bios}} & \multicolumn{3}{c}{\textbf{LongFact}} & \multicolumn{3}{c}{\textbf{WildHallu}} \\
          \cmidrule(lr){3-5} \cmidrule(lr){6-8} \cmidrule(lr){9-11}
          && \multicolumn{1}{c}{$\mathbf{FA}$$\uparrow$} & \multicolumn{1}{c}{$\mathbf{UA}$$\uparrow$} & \multicolumn{1}{c}{$\mathbf{\#Incor}$$\downarrow$} & \multicolumn{1}{c}{$\mathbf{FA}$$\uparrow$} & \multicolumn{1}{c}{$\mathbf{UA}$$\uparrow$} & \multicolumn{1}{c}{$\mathbf{\#Incor}$$\downarrow$} & \multicolumn{1}{c}{$\mathbf{FA}$$\uparrow$} & \multicolumn{1}{c}{$\mathbf{UA}$$\uparrow$} & \multicolumn{1}{c}{$\mathbf{\#Incor}$$\downarrow$} \\
    \midrule
    \multirow{9}{*}{\rotatebox{90}{Mistral-7B-Instruct}}
    &Orig. & 38.8     & --     & 27.9   & 86.2     & --    & 4.55   & 71.5     & --     & 8.31 \\
    & \multicolumn{10}{c}{\textit{Prompt-Based}} \\
    &Unc-Zero & 43.6 & 76.6 & 22.3  & 88.5 & 29.9 & \textbf{1.85} & 75.4 & 48.6 & 5.20 \\
    &Unc-Few & 41.7 & 74.8 & 22.8 & 85.6 & 32.4 & 5.58 & 72.6 & 57.3 & 10.1 \\
    &Pair-Few & 40.3  & 66.7 & 25.9 & 86.1 & 35.1 & 5.08  & 73.4  & 51.6 & 9.80 \\
    &Self-Refine & 38.3  & 57.4 & 27.2 & 86.2 & 42.4 & 4.61  & 72.4  & 50.6 & 8.29 \\
    & \multicolumn{10}{c}{\textit{Training-Based}} \\
    &\sft & 54.5 & 77.1 & 11.4 & 88.6  & 43.5 & 3.21 & 79.2 & 51.1 & 5.73 \\
    &\dpo & \textbf{65.4} & \textbf{80.7} & \textbf{6.54} & \textbf{91.3} & \textbf{54.6} & \underline{2.09} & \textbf{84.4} & \textbf{61.8} & \textbf{3.49} \\
    \midrule
    \multirow{9}{*}{\rotatebox{90}{Llama3-8B-Instruct}}
    &Orig. & 51.9     & --     & 20.4   & 85.5     & --    & 7.45   & 74.4     & --     & 6.24 \\
    & \multicolumn{10}{c}{\textit{Prompt-Based}} \\
    &Unc-Zero & 53.8 & 65.4 & 14.9 & 89.3 & 30.4 & 2.67 & 78.7 & 45.8 & 3.29 \\
    &Unc-Few & 58.7 & 69.1 & 11.7 & 88.3 & 40.6 & 4.01 & 80.2 & 48.6 & 3.44 \\
    &Pair-Few & 60.6 & 46.3 & 9.86  & 86.7 & 42.2 & 4.78 & 84.0 & 48.1 & \textbf{2.33} \\
    &Self-Refine & 51.8 & 40.5 & 18.8 & 85.0 & 27.2 & 6.38 & 77.0 & 28.7 & 5.13 \\
    % \cmidrule(rl){2-11}
    & \multicolumn{10}{c}{\textit{Training-Based}} \\
    &\sft & 58.5 & 69.7 & 11.5 & 87.5  & 44.8 & 4.53 & 78.9 & 44.6 & 5.37 \\
    &\dpo & \textbf{71.4} & \textbf{70.9} & \textbf{4.84} & \textbf{91.5} & \textbf{47.5} & \textbf{2.36} & \textbf{86.0} & \textbf{52.8} & \underline{2.68} \\
    \bottomrule
    \end{tabular}%
    }
  \caption{Comparison of the performance of \sft and \dpo with baseline methods. 
  % We present results on two model scales: Llama3-8B-Instruct and Mistral-7B-Instruct. 
  The best results are \textbf{bolded}, and the second best ones are \underline{underlined}.}
  \label{tab:main}%
\end{table*}%

\subsection{Baselines}
\label{sec:baselines}
In our experiments, we compare our \sft and \dpo\footnote{In our experimental setting, all references to \dpo denote a two-stage training approach combining DPO and SFT. We observed that a model trained solely by DPO does not effectively express uncertainty, which significantly impairs its performance (as shown in Table~\ref{tab:dpo-only}).} against four baseline methods (see implementation details of baselines in Appendix \ref{sec:baseline_prompt}):
\begin{inparaenum}[\it 1)]
    \item \textbf{Unc-Zero}: The model is directly prompted to express uncertainty in its output if it is not sure about any claims.
    \item \textbf{Unc-Few}: Building on Unc-Zero, we provide the model with an additional set of 10 hand-crafted QA examples where uncertainty is explicitly expressed in the answers, following an in-context learning approach. For each example, we provide the question paired with a long-form answer that includes uncertainty expressions. Each example is formatted as \texttt{<Q, $A_{\text{unc}}$>}.
    \item \textbf{Pair-Few}: Extending Unc-Few, we provide the model with both a long-form answer with only certain expressions $A_{\text{cert}}$ and another one with uncertainty expressions $A_{\text{unc}}$ for each query. Each example is formatted as \texttt{<Q, $A_{\text{cert}}$, $A_{\text{unc}}$>}. The goal of including both $A_{\text{cert}}$ and $A_{\text{unc}}$ is to teach the model to differentiate between these two through in-context learning.
    \item \textbf{Self-Refine} \citep{madaan2024self}: Instead of generating the response in one pass, we apply a draft-and-refine setup. The model is asked to first generate an initial response and then refine the uncertain facts into explicit uncertainty expressions in a second pass.
\end{inparaenum}
We also compare our method with several post-hoc methods, where we apply a two-stage pipeline to first estimate the uncertainty for each claim and then revise the most uncertain claims (see Appendix~\ref{sec:post-hoc_baselines} and~\ref{sec:post-hoc_results}).

% Additionally, we used an uncertainty ratio of $\alpha=0.2$ to construct our training data.

\section{Result and Analysis}

\subsection{Main Results}

\rparagraph{LoGU greatly enhances generation performance with higher accuracy and fewer incorrect statements} Table \ref{tab:main} presents our main experimental results. With both models, LoGU-DPO consistently achieves the best performance across all three datasets. Specifically, it improves accuracy from 38.8\% to 65.4\% for Mistral and from 51.9\% to 71.4\% for Llama3 on the \textit{Bios} dataset. This demonstrates that our approach can effectively reduce hallucination and improve overall factuality.

\rparagraph{LoGU-DPO enables the model to accurately express uncertainty} In all baseline methods, although the model can express some uncertainty in long-form generation, the uncertain accuracy remains low. This indicates that the models are merely mimicking uncertainty expressions without reflecting actual uncertainty. While \sft encourages more uncertainty expressions, it does not guarantee that the expressed uncertainty is accurate. \dpo improves this uncertainty misalignment by training the model to express uncertainty only when it may genuinely make mistakes on a given fact, thus achieving better uncertain accuracy. Figure~\ref{fig:case} provides a detailed case study of how \dpo improves over \sft and the original model in generating more accurate uncertainty expressions.

\paragraph{\dpo  shows generalization ability on Out-of-Domain Dataset.}
\label{sec:asqa_result}
To assess the generalizability of our approach, we conduct additional experiments using the ASQA dataset~\citep{stelmakh-etal-2022-asqa}, which focuses on ambiguous factoid questions in long-form question answering. 
Since ASQA does not overlap with the three datasets (\textit{Bios}, \textit{LongFact}, \textit{WildHallu}) used during training, it serves as an ideal out-of-domain benchmark for evaluating our model's robustness. 
As shown in Table~\ref{tab:asqa_result}, our approach outperforms existing methods in ASQA, achieving higher scores in both Fact Accuracy (FA) and Uncertainty Accuracy (UA), further validating its generalizability.

\begin{table}[t!]
\renewcommand{\arraystretch}{1.3} 
\setlength\tabcolsep{2.5pt}
\scalebox{0.92}[0.92]{ 
\small
  \centering
    \begin{tabular}{lvvvddd}
    \toprule
    \multirow{1}{*}{\textbf{Method}} & \multicolumn{3}{c}{\textbf{Mistral-7B}} & \multicolumn{3}{c}{\textbf{Llama3-8B}}
    \\
    \cmidrule(lr){2-4} \cmidrule(lr){5-7}
    & \multicolumn{1}{c}{$\mathbf{FA}$$\uparrow$} & \multicolumn{1}{c}{$\mathbf{UA}$$\uparrow$} & 
    \multicolumn{1}{c}{$\mathbf{\#Incor}$$\uparrow$} & 
    \multicolumn{1}{c}{$\mathbf{FA}$$\uparrow$} & \multicolumn{1}{c}{$\mathbf{UA}$$\uparrow$} & \multicolumn{1}{c}{$\mathbf{\#Incor}$$\uparrow$} \\
    \midrule
    Orig.   & 71.3 & --    & 4.00  & 77.9  & --    & 3.88    \\
    Unc-Zero & 73.4	& 60.7	& 1.73	& 79.3	& 51.7	& \textbf{1.73}  \\
    Self-Refine  & 70.3	& 35.2	& 4.37	& 79.2	& 38.7	& 3.31  \\
    \sft    & 79.9	& 63.2	& 2.49	& 82.4	& 48.8	& 3.17  \\
    \dpo    & \textbf{84.3}	& \textbf{68.9}	& \textbf{1.56}	& \textbf{87.3}	& \textbf{55.2}	& \underline{2.02}
    \\
    \bottomrule
    \end{tabular}
    }
  \caption{Results on Out-of-Domain Dataset (ASQA). \dpo shows robustness when tested on dataset that differ significantly from the training data.}
  \label{tab:asqa_result}
\end{table}

\subsection{Analysis}

\rparagraph{Different Training Data Sources} We explore cases where the training data are collected using the original responses from different source models. In Table \ref{tab:train_other}, we train the Llama3-8B-Instruct model using data constructed from Mistral-7B-Instruct. We have the following two findings: 
\begin{inparaenum}[\it 1)]
\item For \sft, data from another LLM will not always lead to worse performance. This may be because \sft primarily addresses the \textit{uncertainty suppression} issue. As long as the training samples include uncertainty expressions, the model can learn the pattern.
\item For \dpo, using data from another LLM consistently leads to worse performance. We argue that this is because the DPO stage primarily serves to align uncertainty expression with unknown facts, which requires training data that reflects the intrinsic uncertainty of the specific model. Uncertainty for one model may not correspond to uncertainty for another. This finding underscores the importance of using a model's own generations during the DPO stage.
\end{inparaenum}

\rparagraph{Generation Statistics} Table \ref{tab:token_cnt} compares statistics of the original model's generation against the \sft and \dpo outputs. We observe that in \textit{LongFact} and \textit{WildHallu}, the total generation length and the total number of claims remain relatively consistent. The consistency demonstrates that our training strategies do not substantially reduce the amount of generated information. For \textit{Bios}, the average number of tokens decreases considerably, from 400.6 to 255.9. This may be because the original \textit{Bios} responses contain a high number of incorrect claims (27.9 on average), which are reduced after training.
With \dpo, the number of incorrect claims drops significantly to 6.5, leading to much more accurate uncertainty expression. Overall, Table \ref{tab:token_cnt} confirms that our uncertainty training approach effectively preserves the total information content while significantly reducing the number of incorrect claims.


\setlength\tabcolsep{12pt}
\begin{table}[t]
\small
  \centering
    \begin{tabular}{lccc}
    \toprule
    \textbf{Method} & \textbf{FA} & \textbf{UA} & \textbf{\#Incor}\\
    \midrule
    \rowcolor[gray]{0.95}\multicolumn{4}{c}{\textit{Bios}} \\
    \sft & \textbf{58.5}  & \textbf{69.7}  & \textbf{11.5} \\
    \ \ w/ \textit{mismatch data}  &46.8 & 69.6  & 14.4\\ 
    \dpo  & \textbf{71.4}  & \textbf{70.9}  & \textbf{4.84}   \\
    \ \ w/ \textit{mismatch data}  & 69.2  & 68.1 & 5.88 \\ 
    \midrule
    \rowcolor[gray]{0.95}\multicolumn{4}{c}{\textit{LongFact}} \\
    \sft  & \textbf{87.4} & \textbf{44.8} & 4.54 \\
    \ \ w/ \textit{mismatch data} & 86.7  & 43.2 & 3.94 \\ 
    \dpo  & \textbf{91.5} & \textbf{47.5}  & \textbf{2.36} \\
    \ \ w/ \textit{mismatch data} & 89.9  & 44.1 & 3.16 \\ 
    \midrule
    \rowcolor[gray]{0.95}\multicolumn{4}{c}{\textit{WildHallu}} \\
    \sft  & \textbf{78.9} & 44.6 & \textbf{5.36} \\
    \ \ w/ \textit{mismatch data} & 76.5  & 48.2 & 6.30 \\ 
    \dpo  & \textbf{86.0} & \textbf{52.8}  & \textbf{2.68} \\
    \ \ w/ \textit{mismatch data} & 82.7  & 50.6 & 3.39 \\ 
    \bottomrule
    \end{tabular}
  \caption{Results of training Llama3-8B-Instruct with mismatch data (from Mistral-7B-Instruct) show that using the model's own generations in the DPO improves its ability to express uncertainty properly.}
  \label{tab:train_other}
\end{table}

\begin{figure}[t!]
    \centering
    \includegraphics[width=1\columnwidth]{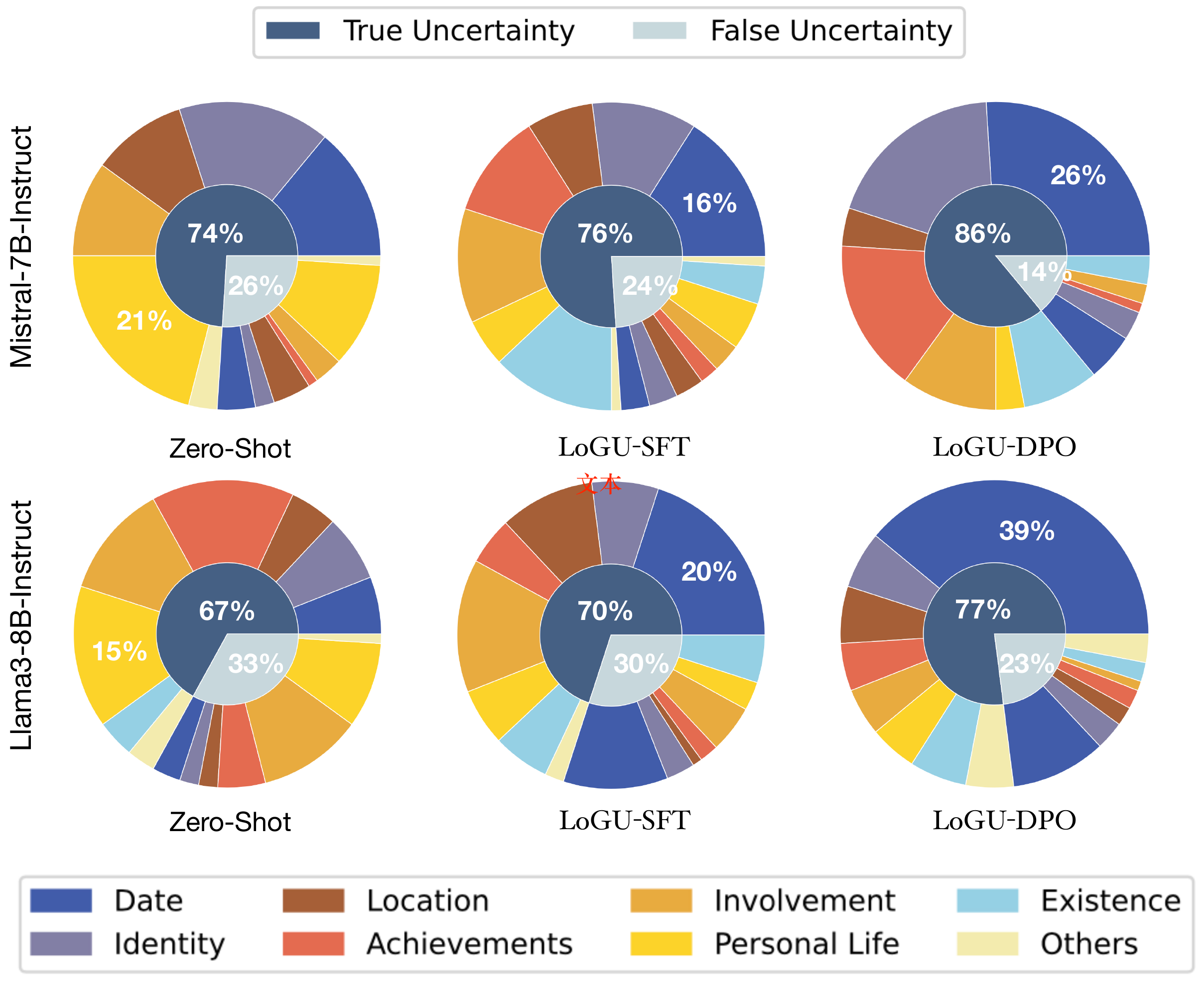}
    \caption{The distribution of uncertainty categories expressed by Zero-Shot, \dpo, \sft on the \textit{Bios} dataset. The inner layer represents the proportion of atomic claims classified as True Uncertainty (\(C_\text{{unc}}^\text{true}\)), while the outer layer shows the distribution across the eight uncertainty categories.}
    \label{fig:category}
\end{figure}


\setlength\tabcolsep{7.2pt}
\begin{table}[t]
\small
  \centering
    \begin{tabular}{lcccc}
    \toprule
    \textbf{Method} & \textbf{\#Cor} & \textbf{\#Incor} 
    & \textbf{\#Unc} 
    & \textbf{\#Token}\\
    \midrule
    \rowcolor[gray]{0.95}\multicolumn{5}{c}{\textit{Bios}} \\
    Orig.  &17.7 & 27.9  & -- & 400.6\\ 
    \sft  &13.6 & 11.4 & 3.82 & 276.8\\ 
    \dpo  &12.3 & 6.54 & 6.32 & 255.9\\ 
    \midrule
    \rowcolor[gray]{0.95}\multicolumn{5}{c}{\textit{LongFact}} \\
    Orig.  &28.4 & 4.55  & -- & 282.6\\ 
    \sft  &24.9 & 3.21  & 3.28 & 283.4\\ 
    \dpo  &21.8 & 2.09  & 5.35 & 277.9\\ 
    \midrule
    \rowcolor[gray]{0.95}\multicolumn{5}{c}{\textit{WildHallu}} \\
    Orig.  &20.8 & 8.31  & -- & 211.9\\ 
    \sft  &21.9 & 5.74  & 2.43 & 268.3\\ 
    \dpo  &18.9 & 3.49  & 4.50 & 266.7\\ 
    \bottomrule
    \end{tabular}
  \caption{Statistics for the original (Orig.), \sft, and \dpo responses using Mistral-7B-Instruct. \#Cor, \#Incor and \#Unc indicate the average number of correct, incorrect and uncertain claims, respectively. \#Token indicates the average token count per response.}
  \label{tab:token_cnt}
\end{table}

\rparagraph{Uncertainty Categories} To gain a deeper understanding of what the model expresses as uncertainty, we conduct an in-depth analysis of uncertainty categories with the \textit{Bios} dataset, as shown in Figure \ref{fig:category}. 
We first manually categorize the topics of uncertain expressions into eight categories that cover all atomic claims, and then use GPT-4o to label each uncertainty expression (details of the categories are provided in Appendix \ref{sec:uncertain_category}).
As illustrated in Figure~\ref{fig:category}, prompt-based methods tend to express true uncertainty in a broad manner. For instance, a significant proportion of uncertain expressions in the personal life category are similar to \texttt{``I have no detailed information about Annabel's early life.''}.
In contrast, training-based methods, particularly \dpo, express uncertainty in a more specific manner, often tied to concrete details, such as \texttt{``I am not sure when Annabel was born.''}. Date-related uncertainty constitutes 26\% in Mistral and 39\% in Llama3, while personal life uncertainty is minimal.
This specificity in expressing uncertainty makes training-based methods more effective in reducing incorrect information about critical details in biographies, particularly dates and timelines, when relevant knowledge is missing.  
However, the general uncertainty expressed by prompt-based methods is less effective in preventing errors in key details, as it often refers to vague pieces of information.
Regarding false uncertainty expressions, both approaches show balanced error distributions across categories, with no category dominating.

\subsection{Human Evaluation}
We conduct a human evaluation to further evaluate the user experience on the uncertainty expression, considering two key aspects: helpfulness and fluency. 
% We defined these dimensions on a scale from 1 to 5. 
\textit{Helpfulness} measures the extent to which the model's output aids the user in understanding the topic and making informed decisions. \textit{Fluency} assesses the cohesion and coherency of the model's output (see detailed information of the human evaluation are provided in Appendix~\ref{sec:human_eval}). We compare the original, Zero-Shot, and \dpo (with uncertainty ratio $\alpha$ values of 0.2 and 0.6) responses.
\begin{figure}[t]
    \centering
    \small
    \definecolor{Red}{RGB}{223, 091, 063}
\definecolor{Green}{RGB}{245, 180, 111}
\definecolor{LightBlue}{RGB}{082, 157, 186}
\definecolor{Blue}{RGB}{053,078,151} 

\pgfplotsset{width=0.95\linewidth,height=0.48\linewidth,compat=1.18}
\footnotesize

\begin{tikzpicture}
    \begin{axis}[
        ybar,
        bar width=8.7pt,
        xtick distance=1,
        symbolic x coords={\pac, \nac},
        xticklabels={a,Helpfulness,Fluency},
        ylabel=Score,
        % xlabel=x values,
        % ylabel=y values,
        % enlarge x limits={abs=0.5},
        ymin=1, ymax=5.4,
        enlarge x limits=0.5,
        scaled ticks=false,
        legend pos=south east,
        % remove the `xticks`
        xtick style={
            /pgfplots/major tick length=0pt,
        },
        legend style={
            nodes={scale=0.6, transform shape},
            legend columns=1,
            font=\footnotesize,
            fill opacity=0.76, % Set the transparency here
        },
       legend image code/.code={
            \draw [#1] (0cm,-0.05cm) rectangle (0.08cm,0.11cm);
            },
    ]
        \addplot+ [
            color=Red!65,
            draw=Red,
            error bars/.cd,
            y dir=both,
            % (changed from `y explicit` so the error bars are (clearly) visible
            y explicit,% relative,
            error mark options={
              Red,
              mark size=0.2pt,
              line width=4pt
            },
            error bar style={line width=0.6pt}
        ] coordinates {
            (\pac,3.0) +- (1.378,1.378)
            (\nac,4.85) +- (0.357,0.357)
        };

        \addplot+ [
            color=Green!65,
            draw=Green,
            error bars/.cd,
            y dir=both,
            % (changed from `y explicit` so the error bars are (clearly) visible
            y explicit, %relative,
            error mark options={
              Green,
              mark size=0.2pt,
              line width=4pt
            },
            error bar style={line width=0.6pt}
        ] coordinates {
            (\pac,3.05) +- (1.023,1.023)
            (\nac,4.1) +- (0.830,0.830)
            % (\aac,79.525)
        };

        \addplot+ [
            color=LightBlue!65,
            draw=LightBlue,
            error bars/.cd,
            y dir=both,
            % (changed from `y explicit` so the error bars are (clearly) visible
            y explicit, % relative,
            error mark options={
              LightBlue,
              mark size=0.2pt,
              line width=4pt
            },
            error bar style={line width=0.6pt}
        ] coordinates {
            (\pac,3.1) +- (1.09,1.09)
            (\nac,4.25) +- (0.94,0.94)
            % (\aac,79.525)
        };
        
        \addplot+ [
            color=Blue!70,
            draw=Blue,
            error bars/.cd,
            y dir=both,
            % (changed from `y explicit` so the error bars are (clearly) visible
            y explicit, % relative,
            error mark options={
              Blue,
              mark size=0.2pt,
              line width=4pt
            },
            error bar style={line width=0.6pt}
        ] coordinates {
            (\pac,3.8) +- (0.67,0.67)
            (\nac,4.45) +- (0.57,0.57)
            % (\aac,79.525)
        };
        
        \legend{
            \text{Origin.},
            \text{Zero-Shot},
            \text{\dpo ($\alpha=0.6$)},
            \text{\dpo ($\alpha=0.2$)},
        }
    \end{axis}
\end{tikzpicture}
    \vspace{-0.2cm}
    \caption{Human evaluation results on helpfulness and fluency of responses generated using different methods.}
    \label{fig:human_eval}
\end{figure}
As shown in Figure \ref{fig:human_eval}, \dpo with a 20\% uncertainty ratio provides the most helpful responses. 
More importantly, annotators reported that using appropriate expressions of uncertainty can highlight information essential to the answers but the model does not know (such as a person's birthdate in a biography), thereby complementing accurate statements.
Although the fluency of the \dpo outputs was slightly lower than that of the original responses, it remained smooth overall. 
Additionally, the variance in both helpfulness and fluency scores for \dpo with a 20\% uncertainty ratio was smaller, indicating that our method produces more stable responses and provides users with a consistently positive experience. 
Moreover, increasing the uncertainty ratio to 60\% results in a noticeable decline in both fluency and helpfulness, suggesting that a higher uncertainty ratio is not always better. More experiments on uncertainty ratio can be found in Appendix~\ref{sec:sft_thresholds}.
%The Zero-Shot method produced the least fluent outputs and struggled to convey important information effectively.

\section{Related Work}

\rparagraph{Fine-tuning LLMs for Factuality}
There has been substantial work on fine-tuning LLMs to improve factuality. \citet{tian2023fine} and \citet{zhang-etal-2024-self} proposed different approaches for annotating responses to fine-tune models using Direct Preference Optimization (DPO). The former focuses on leveraging external knowledge sources, while the latter emphasizes the utilization of the model's own signals to rank the truthfulness of responses. 
Furthermore, \citet{lin2024flame} underscores the importance of factual data in alignment training, employing distinct reward methods for factual and non-factual data to refine model performance. 
Moreover, multiple studies have highlighted that introducing new knowledge during fine-tuning can sometimes lead to increased hallucination \citep{lin2024flame, gekhman2024does, kang2024unfamiliar}. This insight suggests it may be more effective to let models express uncertainty rather than forcing them to state unfamiliar facts.

\rparagraph{Teaching LLMs to Express Uncertainty} 
There are two main approaches to teaching LLMs to express uncertainty: prompt-based and training-based. Prompt-based methods \citep{tian-etal-2023-just, xiong2023llms} design specific prompts to encourage LLMs to convey uncertainty in their responses. On the other hand, training-based approaches \citep{lin2022teaching, li2024know, zhang-etal-2024-r, xu2024sayself} develop specialized datasets to tune the models to express their uncertainty explicitly (\eg "I am sure/unsure"). However, these methods focus on short-form question-answering, where each response contains limited facts. They also provide only one single estimation of uncertainty for the full response. Our work aims to address this gap by enabling models to selectively express uncertainty in responses involving multiple facts for long-form generation.

\rparagraph{Long-form Uncertainty} 
In contrast to expressing uncertainty during generation, several studies focus on externally estimating model uncertainty \citep{zhang2024luq, huang2024calibrating, jiang2024graphbased, fadeeva-etal-2024-fact, shelmanov2025head}. For example, \citet{zhang2024luq} introduced \textsc{Luq}, a consistency-based uncertainty estimation method for long-form generation. \citet{huang2024calibrating} explored the calibration of long-form responses, correlating overall response quality with the associated confidence level. 
However, we argue that post-hoc uncertainty estimation incurs significantly higher computational costs during inference. Therefore, directly expressing uncertainty within the response offers a more efficient approach.
To this end, \citet{band2024linguistic} introduced the concept of linguistic calibration for long-form text generation, which aims to give a calibrated verbalized confidence estimation for each statement.

\section{Conclusion}

In this paper, we introduce the LoGU task, which aims to enhance the ability of LLMs to express uncertainty in long-form generation. To tackle the problems of Uncertainty Suppression and Uncertainty Misalignment, we propose a decomposition-based data collection framework and a two-stage training pipeline with \sft and \dpo. 
Our experiments on three long-form QA datasets demonstrate that our approach significantly improves factual accuracy and uncertain precision, effectively reducing hallucinations without sacrificing the comprehensiveness of the generated content. These results underscore the importance of accurate uncertainty modeling in long-form generation, paving the way for more reliable and trustworthy LLM applications in real-world scenarios.

\section*{Limitation}

This work focuses on enabling models to express uncertainty in factual QA datasets. Future research could explore uncertainty expression in other long-form generation tasks, such as grammar checking, summarization, and translation. 
Regarding factuality metrics, we do not report commonly used metrics for long-form QA tasks, such as ROUGE-L and recall, because the datasets used in our study (\textit{Bios}, \textit{LongFact}, \textit{WildHallu}) do not provide gold-standard answers.

Furthermore, the uncertainty expression in our work is embedded in the text and treated as binary. We do not require models to indicate degrees of uncertainty (\eg "I am 10\% unsure about..."). We argue that binary expression is more human-interpretable and aligns better with natural communication. However, future work could explore machine-interpretable formats, such as attaching confidence tags to each sentence like \citet{zhang2025lovec}.

In Section~\ref{sec:evaluation}, we evaluate uncertainty expression on specific and granular details. Two underexplored areas remain: (1) how to probe known versus unknown knowledge: our current approach uses short-form QA, but more probing methods could be applied; and (2) how to define the specific and granular aspects to focus on; we currently rely on prompting GPT-4o for filtering, but other strategies could be investigated. Our follow-up work \citep{yang2025uncle} further explores these two limitations in detail.

% Our approach is a pioneering effort in expressing uncertainty in long-form generation, and binary uncertainty expression serves as a solid starting point. Meanwhile, we argue that expressing uncertainty in percentages may not be necessary for real-life applications: (1) Risk Indication: Both 10\% and 50\% still convey to users that the statements involve a potential risk of being incorrect. (2) Interpretation Challenges: Users may struggle to interpret specific percentages. For example, the difference between being 60\% certain and 80\% certain can be subtle and subjective, potentially causing confusion or misinterpretation. (3) Cognitive Load: Users might feel overwhelmed by numerical uncertainty, especially in long-form text with frequent instances of uncertainty.

\section*{Ethics Statement}
Our research adheres to rigorous ethical guidelines. We verified the licenses of all softwares and datasets we used in this study and ensured full compliance with their terms. During the human annotation process, all annotators provided informed consent for their data to be used as part of the project. No privacy concerns have been identified. Furthermore, we have thoroughly assessed the project and do not anticipate any additional potential risks.

\section*{Acknowledgement} 
We appreciate the support from the Chinese NSF Major Research Plan, and  (No.92270121). We also acknowledge the use of an icon from Flaticon\footnote{\url{https://www.flaticon.com}} and thank its creators for providing this visually appealing design.

\bibliography{custom,anthology}

\appendix
\onecolumn
\section*{Appendix}

\label{sec:appendix}
\section{Instruction Prompt Examples}
\label{sec:logu_prompt}
The instruction prompts for the revision and assembling procedures in \S\ref{sec:LoGU} are presented in Listing~\ref{listing:prompt}.
%\captionsetup[lstlisting]{position=bottom}
\lstset{
    backgroundcolor=\color[RGB]{245,245,244},
    breaklines=true,
    breakindent=0pt,
    basicstyle=\ttfamily\small,
    %captionpos=b,
    emph={Uncertainty, Revision, Instruction, Refinement, Assembling},
    emphstyle={\bfseries\color{brown}}
}
\begin{lstlisting}[caption={The instruction prompts for Revision and Assembling procedures.},label=listing:prompt]

Revision Instruction:
You will be given a series of atomic facts, each labeled as ##certain## or ##uncertain##. Please adjust each fact following these steps:
- For facts labeled as ##certain##, leave them unchanged.
- For facts labeled as ##uncertain##, adjust them to express uncertainty without focusing on overly specific details. Instead of being uncertain about exact facts, use more general phrases like 'I am not sure when/where/how/what' to convey the uncertainty. 

Here are some uncertainty expressions you can refer to:
- It is uncertain/unclear/not sure/not known
- I am uncertain/unclear/not sure/not known
- There is no information about
- There is no concrete answer about
...

Output each fact(including unchanged facts labeled #ceratin# and adjusted facts with #uncertain# label) in order, as a single line beginning with ###.

For Example:

Facts:
Kang Ji-hwan was born on March 16, 1982. ##uncertain##
Kang Ji-hwan was born in Seoul, South Korea. ##uncertain##
Chief Jones is a respected figure. ##uncertain##
Chief Jones is a respected figure in law enforcement. ##uncertain##
He has had a successful career in the police force. ##uncertain##
He rose through the ranks. ##certain##

Output:
### I do not know when Kang Ji-hwan was born.
### I am not sure where Kang Ji-hwan was born.
### It is unclear whether Chief Jones is a respected figure.
### It is uncertain whether Chief Jones is a respected figure in law enforcement.
### It is uncertain whether he might had a successful career in the police force.
### He rose through the ranks.

Now it's your turn to answer:

Assembling Instruction: 
Your task is to concatenate a provided list of atomic facts, each articulated with either certainty or uncertainty, into a cohesive narrative following the guidelines below:

- All facts in the list, regardless of their certainty, MUST BE included in the generated text. Eliminate any duplicates that may exist.
- Refrain from adding any facts that are not mentioned in the original atomic facts list.
- Your narrative must flow smoothly and coherently. Uncertainty (if present in the atomic facts list) should be expressed diversely, employing phrases such as 'might', 'it is/I am not sure'.

Here is the atomic facts:

{paragraph}

Output:

\end{lstlisting}

\section{Implementation Details.}
\label{sec:implementation_details}
In our experiments, we utilize the Llama3-8B-Instruct~\citep{llama3modelcard} and Mistral-7B-Instruct~\citep{jiang2023mistral} as our base models.

\paragraph{Training Data}
The \sft and \dpo datasets comprise a mix of Bios, WildHallu, and LongFact. Here, we used an uncertainty ratio of $\alpha=0.2$ to construct our training data. 
To enrich the training data and enhance the model's robustness, we implemented data augmentation on the instructions for Bios and WildHallu. We applied five specific transformations as follows:
%\captionsetup[lstlisting]{position=bottom}
\lstset{
    backgroundcolor=\color[RGB]{245,245,244},
    breaklines=true,
    breakindent=0pt,
    basicstyle=\ttfamily\small,
    %captionpos=b,
    emph={Bios, WildHallu, Prompt, Original, Data, Augmentation},
    emphstyle={\bfseries\color{brown}}
}
\begin{lstlisting}[label=training:example]
Original Prompt: 
Bios: Tell me a bio of
WildHallu: In a paragraph, can you tell me about

Data Augmentation:
1. Tell me what you know about
2. Can you provide a detailed introduction of
3. Can you tell me about
4. Can you provide information about

\end{lstlisting}

The specific data set size during the \sft and \dpo phases for Mistral-7B-Instruct and Llama3-8B-Instruct is detailed in Table~\ref{tab:datasize}. We also provide the examples of our \sft and \dpo training data as below:
%\captionsetup[lstlisting]{position=bottom}
\lstset{
    backgroundcolor=\color[RGB]{245,245,244},
    breaklines=true,
    breakindent=0pt,
    basicstyle=\ttfamily\small,
    %captionpos=b,
    emph={Example, LoGU, SFT, DPO, input, output, rejected, chosen, instruction},
    emphstyle={\bfseries\color{brown}}
}
\begin{lstlisting}[label=sft:example]
LoGU-SFT Example:
{
    "instruction": "Tell me a bio of Suthida.",
    "input": "",
    "output": "Suthida Vajiralprabha is the current queen consort of Thailand. Before her ascension to this prestigious role, she was born into a middle-class family and rose through the ranks in the Royal Thai Police, where she was known for her dedication and hard work. Although it is uncertain, some say she might have worked as a nurse, but I am not sure where she might have done so in Bangkok.\n\nIn a surprising turn of events, Suthida Vajiralprabha married King Maha Vajiralongkorn of Thailand on May 1, 2019. Following her marriage, she was given the title of Queen Consort and has since taken on various royal duties, often seen attending official events with the king and engaging in charitable work.\n\nDespite her royal status, Suthida Vajiralprabha is known for her simple and down-to-earth personality. Her humility and dedication to serving the people have won the hearts of many Thais. While I do not know what her specific royal title is, I can affirm that she is fluent in Thai and continues to be a beloved figure in Thailand."
},
LoGU-DPO Example:
{
    "instruction": "Who is Gottfried Wilhelm Leibniz?",
    "input": "",
    "chosen": "Gottfried Wilhelm Leibniz, born on July 1, 1646, and passing away on November 14, 1716, was a German philosopher, mathematician, and polymath. Widely regarded as one of the greatest thinkers of the late 17th century, Leibniz made significant contributions to various fields including metaphysics, logic, theology, and particularly mathematics. Notably, he developed calculus independently of Sir Isaac Newton and invented the Leibniz wheel, although I am not sure whether he also invented the Leibniz calculus.\n\nLeibniz is perhaps most famous for his philosophical concept of monadology, which posits that the universe is made up of individual, simple substances called monads. While he was undeniably a prolific writer, the exact number of letters and manuscripts he produced remains uncertain. \n\nAs a member of several European academies, Leibniz held various positions of influence throughout Europe. However, it is not clear whether he made significant contributions to the fields of physics and diplomacy. Overall, Leibniz's legacy as a thinker who traversed multiple disciplines has ensured his lasting impact on the intellectual landscape.",
    "rejected": "Gottfried Wilhelm Leibniz (July 1, 1646 \u2013 November 14, 1716) was a German philosopher, mathematician, and polymath who is considered one of the greatest thinkers of the late 17th century. He is best known for his contributions to philosophy, particularly in the areas of metaphysics, logic, and theology. Leibniz is also known for his work in mathematics, including his development of the calculus independent of Sir Isaac Newton, and his invention of the Leibniz wheel and the Leibniz calculus.\n\nLeibniz is perhaps most famous for his philosophical concept of monadology, which posits that the universe is made up of individual, simple substances called monads. He also made significant contributions to the fields of physics, law, and diplomacy. Leibniz was a prolific writer and is believed to have produced over 4,000 letters and manuscripts during his lifetime. He was a member of several European academies and held various positions of power and influence throughout Europe."
}

\end{lstlisting}
\begin{table}[ht]
    \centering
    \setlength\tabcolsep{8pt}
    \begin{tabular}{lcccc}
    \toprule
    Stage & \textit{bios} & \textit{LongFact} &\textit{WildHallu} &Total \\
    \midrule
    \rowcolor[gray]{0.95}\multicolumn{5}{c}{\textit{Mistral-7B-Instruct}} \\
    \sft & 7970 & 29160 &1548 &38678 \\
    \dpo & 3555 & 19096 &911 &23562 \\
    \rowcolor[gray]{0.95}\multicolumn{5}{c}{\textit{Llama3-8B-Instruct}} \\
    \sft & 8475 &27365 & 1417 &37257 \\
    \dpo & 3933  & 19096 &851 &23880 \\
    \bottomrule
    \end{tabular}
    \caption{Training dataset size for \sft and \dpo.}
    \label{tab:datasize}
\end{table}

\paragraph{Fine-tuning Details}
We run \dpo and \sft experiments with 8 NVIDIA A100-40GB GPUs. 
We conduct experiments with the LlamaFactory code base\footnote{https://github.com/hiyouga/LLaMA-Factory}.
During the \sft phase, we fine-tune the base models on a mixed dataset of 40k. 
In the \dpo phase, we continue fine-tuning the models on a mixed dataset of 20k instances.
Building upon prior research, which highlights the MLP layer as a crucial element for embedding knowledge within the LLM transformer architecture \citep{de2021editing}, we
only fine-tune the weight matrix of the attention
layer using LoRA~\citep{Hu2021LoRALA}. 
This method allows us to adjust the model's ability to express knowledge boundaries without altering its internal knowledge structure. 
The configurations of our hyper-parameters are detailed in Table~\ref{tab:configuration}.

\begin{table}[ht]
    \centering
    \setlength\tabcolsep{4pt}
    \begin{tabular}{lcc}
    \toprule
    Configuration & \sft & \dpo \\
    \midrule
    Model & Mistral-7B(Llama3-8B)-Instruct  & Mistral-7B(Llama3-8B)-Instruct \\
    Number of epochs & 3 & 3 \\
    % Devices &8 A100 GPU $(40 \mathrm{~GB})$ &8 A100 GPU $(40 \mathrm{~GB})$\\
    Total Batch size & 64 samples & 64 samples \\
    Optimizer & Adam~\cite{kingma2014adam} & Adam~\cite{kingma2014adam} \\
    & $(\beta_1=0.9, \beta_2=0.98, \epsilon=1 \times 10^{-8})$ & $(\beta_1=0.9, \beta_2=0.98, \epsilon=1 \times 10^{-8})$\\
    Learning rate & $5 \times 10^{-5}$ & $1 \times 10^{-5}$ \\
    Warmup Ratio & 0.1 & 0.1 \\
    LoRA Target & $\mathrm{q}_{\text{proj}}, \mathrm{v}_{\text{proj}}$ & $\mathrm{q}_{\text{proj}}, \mathrm{v}_{\text{proj}}$\\
    LoRA Parameters & $r=8, \alpha=16, \text{dropout}=0.05$ & $r=8, \alpha=16, \text{dropout}=0.05$ \\
    Training Time &1h 31m 30s (1h 43m 3s) &1h 55m (1h 3m 8s) \\
    \bottomrule
    \end{tabular}
    \caption{Fine-tuning hyper-parameters for \sft and \dpo.}
    \label{tab:configuration}
\end{table}

\paragraph{Evaluation}
We use vLLM \citep{kwon2023efficient} for our LLM inference tasks, with the following parameters: temperature = 0.7, top-$p$ = 0.95, and a maximum output of 1024 tokens. 
For Fact-Cheking, Uncertainty Revision, Assemble procedures, we set the temperature to 0. 
We use GPT-4o as the auxiliary model for generating atomic claims and fact-checking the LLM.

\clearpage
\section{Baselines}
\subsection{Generative Methods}
\label{sec:baseline_prompt}
Here we list the prompts of our baselines (zero, few-shot, paired few-shot and self-refine) in $\mathsection$\ref{sec:baselines}.
%\captionsetup[lstlisting]{position=bottom}
\lstset{
    backgroundcolor=\color[RGB]{245,245,244},
    breaklines=true,
    breakindent=0pt,
    basicstyle=\ttfamily\small,
    %captionpos=b,
    emph={Zero,-,Shot,(,Bios,), LongFact, WildHallu, Few, Examples, Instruction, Paired, Question, Answer, Good, Bad, Refine, Self},
    emphstyle={\bfseries\color{brown}}
}
\begin{lstlisting}[caption={The instruction prompts of key procedures.},label=listing:baselines]
Zero Shot(Bios):
Your task is to write a biography for a specific entity. You should express uncertainty for any information you are not familiar with.

Zero Shot(LongFact):
Your task is to answer the given question about a specific object (e.g., a person, place, event, company, etc.). Express uncertainty for any information you are not familiar with (e.g., 'I am not sure if/whether', 'It is uncertain that').

Zero Shot(WildHallu):
Your task is to write a paragraph for a specific entity. You should express uncertainty for any information you are not familiar with (e.g., 'I am not sure if/whether', 'It is uncertain that').

Few Shot Examples:
Question: In a paragraph, could you tell me what you know about Vycta.co.ke?
Answer: Vycta.co.ke is a Kenyan-based digital health platform that provides healthcare services to patients. Additionally, Vycta.co.ke offers various health-related services, ensuring comprehensive support for its users.

Paired Few Shot Examples:
Question: In a paragraph, could you tell me what you know about Vycta.co.ke?
Good Answer: Vycta.co.ke is a Kenyan-based digital health platform that provides healthcare services to patients. Additionally, Vycta.co.ke offers various health-related services, ensuring comprehensive support for its users.
Bad Answer: Vycta.co.ke is a Kenyan-based digital health platform that provides a range of healthcare services to patients and healthcare providers. The platform offers a telemedicine service that allows patients to consult with doctors remotely, as well as a health insurance platform that provides access to affordable healthcare coverage. Vycta also offers a range of health-related products and services, including a health monitoring app that allows patients to track their health metrics and receive personalized health advice. The platform aims to increase access to healthcare services, particularly in rural and underserved areas, and to make healthcare more affordable and convenient for patients.

Self Refine:
Revise your previous response to the question by adjusting the expression of certainty. 
For any information you are unsure about, frame your response to reflect this uncertainty. Keep the expression of information you are certain about unchanged.

For example: 
Question: Tell me a bio of Miguel Angel Felix Gallardo.
Answer: Miguel Angel Felix Gallardo, also known as "El Padrino" (The Godfather), is a Mexican drug lord and former leader of the Guadalajara cartel, one of the most powerful drug trafficking organizations in Mexico during the 1980s. He was born on August 7, 1946, in La Barranca de Agua Caliente, Sinaloa, Mexico.
Gallardo began his criminal career in the 1960s, working as a small-time smuggler and drug dealer. He quickly rose through the ranks, becoming a key player in the Mexican drug trade and eventually forming his own organization, the Guadalajara cartel, in the 1970s.
Refine: Miguel Angel Felix Gallardo, who is widely known as 'El Padrino' (The Godfather), is a notorious Mexican drug lord. He emerged as a prominent figure in the criminal underworld by forming the Guadalajara cartel in the 1970s. Gallardo's rise to power was marked by his adept maneuvering within the Mexican drug trade, where he quickly became a key player.

Now it is your turn to refine:
Question: {question}
Answer: {answer}
Refine:
\end{lstlisting}

\subsection{Post-hoc Methods}
\label{sec:post-hoc_baselines}
We incorporate two baselines—P(true) ~\citep{kadavath2022languagemodelsmostlyknow} and Semantic Entropy (SE) ~\citep{kuhn2023semanticuncertaintylinguisticinvariances}—which rely on post-hoc uncertainty estimates derived from the model's internal signals. 
For each baseline, we compute a post-hoc uncertainty score (P(true) or SE) for each atomic claim. 
Based on these scores, we rank the atomic claims within each dataset and set a threshold for uncertain claims (we chose the bottom 10\%). 
These claims were then modified using GPT-4 to explicitly express uncertainty.

\subsection{Comparision of \sft, \dpo and  Post-hoc Methods.}
\label{sec:post-hoc_results}
In this section, we compare the post-hoc baselines discussed in \S\ref{sec:post-hoc_baselines} with our \sft and \dpo methods. 
Similar to the main experiments, we evaluate on three datasets: \textit{Bios}, \textit{LongFact}, and \textit{WildHallu}, using Mistral-7B as the backbone model. The evaluation metrics are Factual Accuracy (FA) and Uncertainty Accuracy (UA). 
As shown in Table~\ref{tab:post-hoc}, our LoGU-DPO outperforms the post-hoc uncertainty baselines (P(true) and SE) in both FA and UA across all three datasets. We hypothesize that this ineffectiveness stems from the accuracy limitations of these uncertainty estimation methods in the context of long-text generation.
\begin{table*}[ht!]
\setlength\tabcolsep{12pt}
%\scalebox{0.95}[0.95]{ 
\small
  \centering
    \begin{tabular}{lcccccc}
    \toprule
    \multirow{2}{*}{\textbf{Method}} & \multicolumn{2}{c}{\textbf{Bios}} & \multicolumn{2}{c}{\textbf{LongFact}} & \multicolumn{2}{c}{\textbf{WildHallu}} \\
    \cmidrule(lr){2-3} \cmidrule(lr){4-5} \cmidrule(lr){6-7}
    & \multicolumn{1}{c}{FA$\uparrow$} & \multicolumn{1}{c}{UA$\uparrow$} & \multicolumn{1}{c}{FA$\uparrow$} & \multicolumn{1}{c}{UA$\uparrow$} & \multicolumn{1}{c}{FA$\uparrow$} & \multicolumn{1}{c}{UA$\uparrow$} \\
    \midrule
    Orig.   & 38.8 & --    & 86.2 & --    & 71.5 & --    \\
    P(true) & 45.1 & 72.6  & 87.4 & 26.8  & 73.8 & 44.2  \\
    SE      & 41.7 & 83.1  & 88.4 & 36.3  & 73.4 & 53.1  \\
    \sft    & 54.5 & 77.1  & 88.6 & 43.5  & 79.2 & 51.1  \\
    \dpo    & 65.4 & \textbf{80.7} & \textbf{91.3} & \textbf{54.6} & 84.4 & \textbf{61.8} \\
    \bottomrule
    \end{tabular}
  \caption{Comparison of \sft, \dpo and Post-hoc Methods.}
  \label{tab:post-hoc}
\end{table*}

\section{Comparision of \dpo and DPO-Only.}
In this section, we compare the performance of \dpo and DPO-Only. 
Similar to the main experiments, we evaluate the models on three datasets: \textit{Bios}, \textit{LongFact}, and \textit{WildHallu}, using Factual Accuracy (FA) and Uncertainty Accuracy (UA) as metrics. 
As shown in Table~\ref{tab:dpo-only}, we find that \dpo achieves the best performance, while training with DPO alone significantly decreases model performance, with UA metrics even falling behind \sft.
\begin{table*}[ht!]
\setlength\tabcolsep{10pt}
%\scalebox{0.95}[0.95]{ 
\small
  \centering
    \begin{tabular}{llcccccc}
    \toprule
     & \multirow{2}{*}{\textbf{Method}} & \multicolumn{2}{c}{\textbf{Bios}} & \multicolumn{2}{c}{\textbf{LongFact}} & \multicolumn{2}{c}{\textbf{WildHallu}} \\
          \cmidrule(lr){3-4} \cmidrule(lr){5-6} \cmidrule(lr){7-8}
          & 
          & \multicolumn{1}{c}{FA$\uparrow$} & \multicolumn{1}{c}{UA$\uparrow$} & \multicolumn{1}{c}{FA$\uparrow$} & \multicolumn{1}{c}{UA$\uparrow$} & \multicolumn{1}{c}{FA$\uparrow$} & \multicolumn{1}{c}{UA$\uparrow$} \\
    \midrule
    \multirow{4}{*}{\rotatebox{90}{Mistral-7B}}
    &Orig. & 38.8     & --     & 86.2     & --    & 71.5     & --     \\
    &\sft & 54.5 & 77.1 & 88.6  & 43.5 & 79.2 & 51.1 \\
    &\dpo & 65.4 & \textbf{80.7} & \textbf{91.3} & \textbf{54.6}  &84.4 & \textbf{61.8} \\
    & DPO-only & \textbf{68.7} &76.6 & 91.8 & 46.8 & 83.4 & 52.2 \\
    \midrule
    \multirow{4}{*}{\rotatebox{90}{Llama3-8B}}
    &Orig. & 51.9     & --    & 85.5     & --    & 74.4     & --  \\
    &\sft & 58.5 & 69.7 & 87.5  & 44.8 & 78.9 & 44.6 \\
    &\dpo & 71.4 & \textbf{70.9} & \textbf{91.5} & \textbf{47.5} & \textbf{86.0} & \textbf{52.8} \\
    & DPO-only & \textbf{74.8} &62.4 & 90.1 & 42.1 & 83.6 & 46.7 \\
    \bottomrule
    \end{tabular}%
%    }
  \caption{Comparision of \dpo and DPO-Only.}
  \label{tab:dpo-only}%
\end{table*}%

\section{Impact of the Uncertainty Ratio Parameter $\alpha$.}
\label{sec:sft_thresholds}
To investigate the impact of the uncertainty ratio ($\alpha$) in the training data on the final results, we train the Mistral-7B-Instruct using training data with different values of $\alpha$ during the SFT stage. As shown in Table~\ref{tab:sft-thresholds}, the results indicate that $\alpha = 0.2$ yields the highest Fact Accuracy (FA) and Uncertainty Accuracy (UA) across the three datasets, highlighting a trade-off between uncertainty expression and response quality.

\begin{table*}[ht!]
\setlength\tabcolsep{12pt}
%\scalebox{0.95}[0.95]{ 
\small
  \centering
    \begin{tabular}{lcccccc}
    \toprule
    \multirow{2}{*}{\textbf{Method}} & \multicolumn{2}{c}{\textbf{Bios}} & \multicolumn{2}{c}{\textbf{LongFact}} & \multicolumn{2}{c}{\textbf{WildHallu}} \\
    \cmidrule(lr){2-3} \cmidrule(lr){4-5} \cmidrule(lr){6-7}
    & \multicolumn{1}{c}{FA$\uparrow$} & \multicolumn{1}{c}{UA$\uparrow$} & \multicolumn{1}{c}{FA$\uparrow$} & \multicolumn{1}{c}{UA$\uparrow$} & \multicolumn{1}{c}{FA$\uparrow$} & \multicolumn{1}{c}{UA$\uparrow$} \\
    \midrule
    \sft($\alpha=0.2$)   & 54.5	& 77.1    & 88.6	& 43.5  & 79.2 & 51.1   \\
    \sft($\alpha=0.4$) & 51.8	& 74.0	 & 87.4	& 42.6 & 79.7& 46.0  \\
    \sft($\alpha=0.6$)      & 52.9	& 71.6  & 87.3 & 43.6  & 78.4	 & 48.4  \\
    \sft($\alpha=0.8$)    & 47.0 & 73.7  & 87.5	& 42.3 & 78.3 & 49.1
  \\
    \bottomrule
    \end{tabular}
  \caption{Comparison of \sft, \dpo and Post-hoc Methods.}
  \label{tab:sft-thresholds}
\end{table*}

\section{Case Study}
We randomly selected a sample from the \textit{WildHallu} dataset to compare Original Answer and \dpo Answer. Each atomic claim in the generated responses was manually fact-checked using Wikipedia and other sources accessible via Google. Correct statements were highlighted in green, incorrect statements in red, and uncertainty expressions in blue.

\begin{figure}[t]
    \centering
    \includegraphics[width=1\columnwidth]{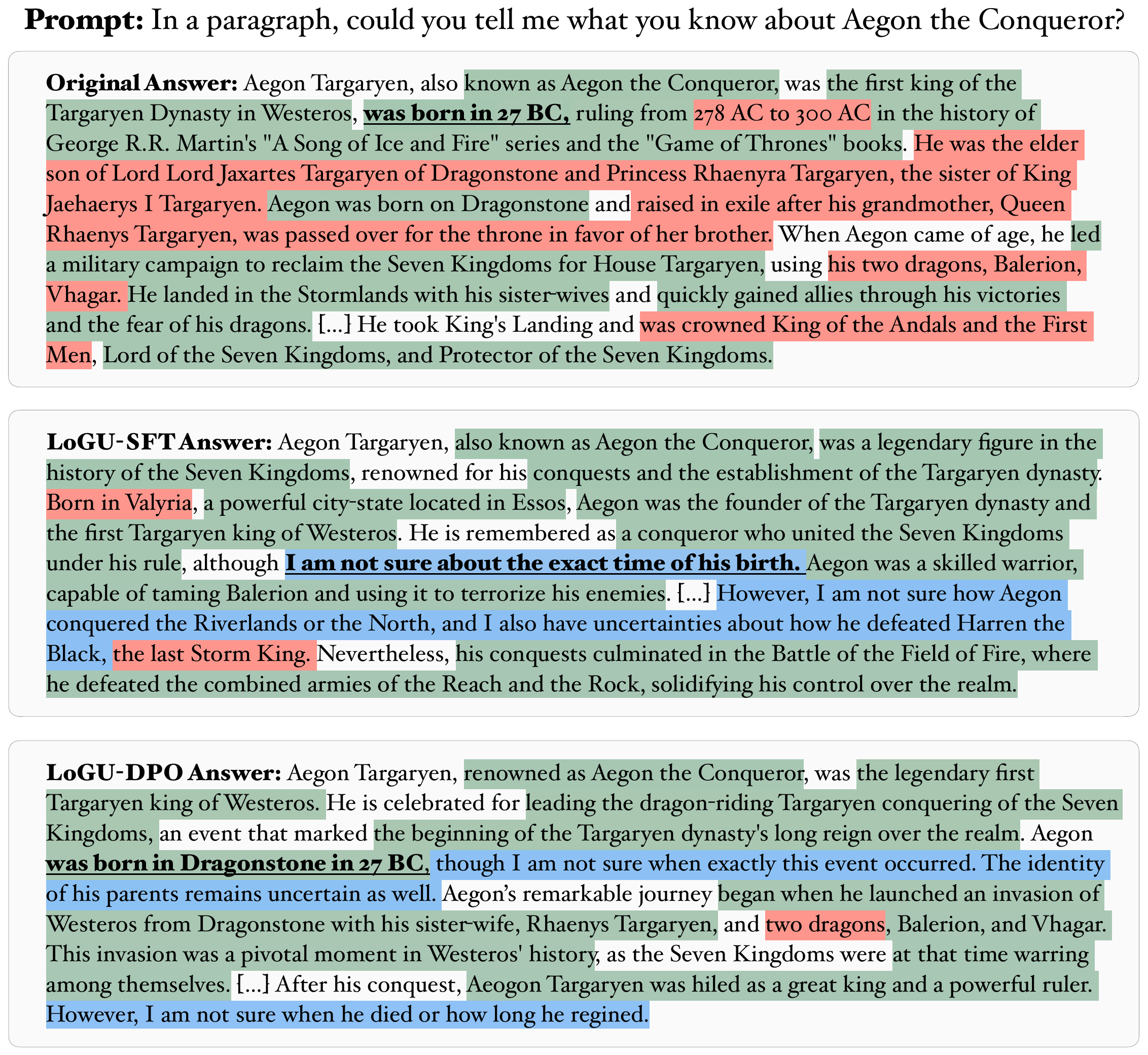}
    \caption{Case Study: Green indicates factually correct statements, red represents incorrect ones, and blue highlights uncertain expressions. The Original Answers includes a mix of correct and incorrect statements, while the \dpo answers retains the correct statements and revises the incorrect ones into expressions of uncertainty.
}
    \label{fig:case}
\end{figure}

As shown in Figure~\ref{fig:case}, the original answer contains a mix of correct and incorrect statements, which diminishes the overall factual accuracy of the generated content. In the case of \sft, it significantly reduces the number of incorrect statements. However, it also introduces uncertainty in the expression of some correct facts. For example, in the original answer, Aegon the Conqueror's birth time is correctly stated as \texttt{``27BC''}, but in the \sft answer, this fact is expressed with uncertainty. In contrast, the \dpo response preserves the correct statements while replacing incorrect ones with expressions of uncertainty. For instance, the original answer incorrectly identifies Aegon the Conqueror's parents as \texttt{``He was the elder son of Lord Jaxartes Targaryen of Dragonstone and Princess Rhaenyra Targaryen.''} However, the \dpo response addresses this uncertainty by stating, \texttt{``The identity of his parents remains uncertain.''}, thereby significantly improving the overall accuracy of the response.

\clearpage
\section{Uncertainty Categories}
Detailed descriptions of eight categories derived from uncertainty in atomic claims.
\label{sec:uncertain_category}
%\captionsetup[lstlisting]{position=bottom}
\lstset{
    backgroundcolor=\color[RGB]{245,245,244},
    breaklines=true,
    breakindent=0pt,
    basicstyle=\ttfamily\small,
    %captionpos=b,
    emph={Date, Timing, Uncertainty, Identity, and, Occupation, Location, Geography, Achievements, Contributions, Involvement, Action, Personal, Life, Background, Information, Existence, Factual, Veracity, Others, Example},
    emphstyle={\bfseries\color{brown}}
}
\begin{lstlisting}[label=category:prompt]

0: Date and Timing Uncertainty.
    This label applies to uncertainties about the specific times of events, important dates in a person's life, or any time-related facts.
    Example: "When Marie Alexandrine Becker received the Lasker-DeBakey Clinical Medical Research Award is [/uncertain/]." or "The timing of when Bella Akhmadulina was awarded the Lenin Komsomol Prize for Literature is [/uncertain/]."

1: Identity and Occupation Uncertainty.
    Use this for uncertainties about the real existence or the professional role of individuals.
    Example: "It is [/uncertain/] whether Chief Jones is a real person." or "Whether Ravi is a fellow of the IACR is [/uncertain/]."

2: Location and Geography Uncertainty.
    This covers uncertainties related to the places of birth, education, residence, or any geographical locations.
    Example: "The place where Sara Paxton was born is [/uncertain/]." or "The place where Lousteau earned his undergraduate degree is [/uncertain/]."

3: Achievements and Contributions Uncertainty.
    This involves uncertainties about someone's achievements, awards, or any professional contributions.
    Example: "Whether Antonio Gasalla received the Premio Regione Piemonte is [/uncertain/]." or "The awards won by Wilfredo Vargas is [/uncertain/]."

4: Involvement and Action Uncertainty.
    Pertains to uncertainties about someone's participation in projects, campaigns, or actions.
    Example: "Liam Payne's involvement with the 'Text Santa' campaign is [/uncertain/]." or "It's unclear whether Mauro Icardi was loaned to Sampdoria."

5: Personal Life and Background Information Uncertainty.
    Use this for uncertainties about personal details, family background, or other private aspects.
    Example: "Kourosh Zolani's personal life is [/uncertain/]." or "Many aspects of Rivera's early life are [/uncertain/]."

6: Existence and Factual Veracity Uncertainty.
    This label is for uncertainties about the factual existence of events, appearances in media, or historical facts.
    Example: "It is [/uncertain/] whether Marianne McAndrew has appeared in 'Law & Order: Special Victims Unit: The Movie.'" or "It is uncertain whether Virginia Valli appeared in 'The Big Combo'."
    
7: Others.
    Anything that does not fit into the categories above.
    \end{lstlisting}

\clearpage
\section{Human Evaluation}
\label{sec:human_eval}
We randomly selected 100 queries from the \textit{bios}, \textit{WildHallu}, and \textit{LongFact} datasets. 
Human annotators were instructed to assess the usefulness and fluency of the responses generated by Mistral-7B using different methods. We defined these dimensions on a scale from 1 to 5.
We recruited three students with bachelor’s degrees in computer science and fluent English to conduct the annotations. These annotators were not involved in our project and had no prior discussions related to it.
To measure inter-annotator agreement, we used Fleiss’ Kappa, obtaining a score of 0.759, which indicates substantial agreement ~\citep[approaching the “almost perfect" range of 0.8-1.0;][]{landis1977measurement}. The annotators were compensated above the local minimum hourly wage.
The instruction to guide human annotators to evaluate the \textbf{Helpfulness} and \textbf{Fluency} of Long-Form Responses are present in the listing~\ref{listing:helpfulness_eval} and ~\ref{listing:fluency_eval}.
%\captionsetup[lstlisting]{position=bottom}
\lstset{
    backgroundcolor=\color[RGB]{245,245,244},
    breaklines=true,
    breakindent=0pt,
    basicstyle=\ttfamily\small,
    %captionpos=b,
    emph={Definition, Correct, Incorrect, Uncertain, Helpfulness, Information, Rating, Scale, Not, Minimally, Moderately, Very, Of, Helpful},
    emphstyle={\bfseries\color{brown}}
}
\begin{lstlisting}[caption={Instructions for Human Annotators: Evaluating the Helpfulness of Long-Form Responses
},label=listing:helpfulness_eval]
Your task is to evaluate the helpfulness of each long-form response based on the proportion of Correct, Incorrect, and Uncertain information. Each long-form response will be highlighted to indicate these categories for different atomic claims. Your goal is to assess how these proportions impact the user's ability to understand the topic and make informed decisions.

### Definition Of Helpfulness:
Helpfulness refers to how effectively the response provides **useful, accurate, and relevant** information. A long-form response typically contains multiple atomic claims that can be categorized into:

- Correct Information: Accurate and relevant facts or details that directly address the query. The more correct information a response contains, the more useful and reliable it is for the user.

- Incorrect Information: Misleading, inaccurate, or irrelevant details that detract from the response's usefulness. Incorrect information can confuse the user or lead to misunderstandings, which reduces the overall helpfulness.

- Uncertain Information: Statements that express ambiguity, limited knowledge, or lack of confidence. Uncertainty can be Helpful if it highlights the limits of knowledge or provides context, and it is generally better than giving incorrect information for critical points. However, if it is vague, unexplained, or excessive, it may reduce the clarity and impact of the response.

### Rating Scale:
You will rate the helpfulness of each long-form response on a scale from 1 to 5, depending on the balance of Correct, Incorrect, and Uncertain Information:

#### 1: Not Helpful
- The response is dominated by incorrect information or excessive, vague uncertainty, making it mostly misleading or confusing. Correct information, if present, is minimal and not useful.

#### 2: Minimally Helpful
- The response contains some correct information, but it is heavily outweighed by significant incorrect claims or unclear uncertainty. While there are elements of usefulness, the overall response may confuse or mislead the user.

#### 3: Moderately Helpful
- The response presents a mixture of correct, uncertain, and incorrect information. The majority of the response is correct and relevant, but incorrect or unclear claims detract from its overall helpfulness.

#### 4: Helpful
- The response is mostly correct, with only minor incorrect details or justified uncertainty. Uncertainty is presented clearly and contributes to the user's understanding. The correct information outweighs any minor errors or vagueness, making the response helpful overall.
  
#### 5: Very Helpful
- The response is highly informative, with a high proportion of correct information and well-justified uncertainty where appropriate. Incorrect information is minimal or absent. Even if uncertainty is present, it enhances understanding and complements the value of the correct information.
  
\end{lstlisting}
%\captionsetup[lstlisting]{position=bottom}
\lstset{
    backgroundcolor=\color[RGB]{245,245,244},
    breaklines=true,
    breakindent=0pt,
    basicstyle=\ttfamily\small,
    %captionpos=b,
    emph={Definition, Fluency, Rating, Scale, Not, Fluent, Minimally, Moderately, Very, Of},
    emphstyle={\bfseries\color{brown}}
}
\begin{lstlisting}[caption={Instructions for Human Annotators: Evaluating the Fluency of Long-Form Responses
},label=listing:fluency_eval]
Your task is to evaluate the fluency of each long-form response. Fluency refers to the readability, flow, and naturalness of the language used in the response. You should assess the fluency based on overall how smoothly the response reads and whether it is free from awkward phrasing or unnatural language. You should not only judge the fluency purely on the presence of uncertain expressions. 

### Definition Of Fluency:
Fluency refers to how naturally and clearly the response is written. A fluent response should be well-structured, easy to read, and free of grammatical errors or awkward phrasing. Fluency is important in maintaining the reader's engagement and ensuring the response is easily understood.


### Rating Scale:
You will rate the fluency of each long-form response on a scale from **1 to 5**, where the score reflects how smoothly and naturally the response reads.

#### 1: Not Fluent
- The response is difficult to read due to frequent awkward phrasing, or disjointed structure. It requires significant effort to follow, and the meaning may be unclear.

#### 2: Minimally Fluent
- The response is somewhat readable, but it contains frequent awkward phrasing that disrupt the flow. The meaning is mostly clear, but the language is clunky and requires effort to process.

#### 3: Moderately Fluent
- The response is generally fluent, but there are some instances of awkward phrasing. The meaning is clear, but the overall readability could be improved.

#### 4: Fluent
- The response is well-written, with only minor instances of slightly awkward phrasing. The response flows naturally, and the meaning is clear without any interruptions to the reader's understanding.

#### 5: Very Fluent
- The response reads very smoothly, with no awkward phrasing. It is natural, clear, and well-structured, making it very easy for the reader to follow.
\end{lstlisting}

\end{document}